\documentclass[sigconf]{acmart}
\usepackage{multirow}
\usepackage{tabularx}

\graphicspath{{./Images/}{./Results/}}
\hypersetup{
    colorlinks = true,
    linkcolor={blue}
}
\setcopyright{rightsretained}

\acmArticle{30}
\acmPrice{15.00}
\acmDOI{10.1145/3571600.3571630}
\acmISBN{978-1-4503-9822-0/22/12}
\acmPrice{15.00}
\editor{Soma Biswas}
\editor{Shanmuganathan Raman}
\editor{Amit K Roy-Chowdhury}

\acmConference[ICVGIP'22]{13th Indian Conference on Computer Vision, Graphics and Image Processing}{December 2022}{Gandhinagar, India}
\acmYear{2022}
\copyrightyear{2022}

\acmPrice{15.00}
\setcopyright{acmcopyright}

\begin{document}
\acmBooktitle{13th Indian Conference on Computer Vision, Graphics and Image Processing}
\title{Towards Realistic Underwater Dataset Generation and Color Restoration}
\titlenote{Produces the permission block, and
  copyright information}

\author{Neham Jain}
\affiliation{%
  \institution{IIT Madras}
  \city{Chennai}
  \state{Tamil Nadu}
  \country{India}
  \postcode{600036}
}
\email{nehamjain2002@gmail.com}
\author{Gopi Raju Matta}
\affiliation{%
  \institution{IIT Madras}
  \city{Chennai}
  \state{Tamil Nadu}
  \country{India}
  \postcode{600036}
}
\email{ee17d021@smail.iitm.ac.in}

\author{Kaushik Mitra}
\affiliation{%
  \institution{IIT Madras}
  \city{Chennai}
  \state{Tamil Nadu}
  \country{India}
  \postcode{600036}
}
\email{kmitra@ee.iitm.ac.in}

\renewcommand{\shortauthors}{}

\begin{abstract}
Recovery of true color from underwater images is an ill-posed problem. This is because the wide-band attenuation coefficients for the RGB color channels depend on object range, reflectance, etc. which are difficult to model. Also, there is backscattering due to suspended particles in water. Thus, most existing deep-learning based color restoration methods, which are trained on synthetic underwater datasets, do not perform well on real underwater data. This can be attributed to the fact that synthetic data cannot accurately represent real conditions. To address this issue, we use an image to image translation network to bridge the gap between the synthetic and real domains by translating images from synthetic underwater domain to real underwater domain.  Using this multimodal domain adaptation technique, we create a dataset that can capture a diverse array of underwater conditions. We then train a simple but effective CNN based network on our domain adapted dataset to perform color restoration. Code and pre-trained models can be accessed at \href{https://github.com/nehamjain10/TRUDGCR}{https://github.com/nehamjain10/TRUDGCR}
\end{abstract}

%
%



\maketitle

\begin{figure}[!htp]
    \centering
    \includegraphics[width=\columnwidth]{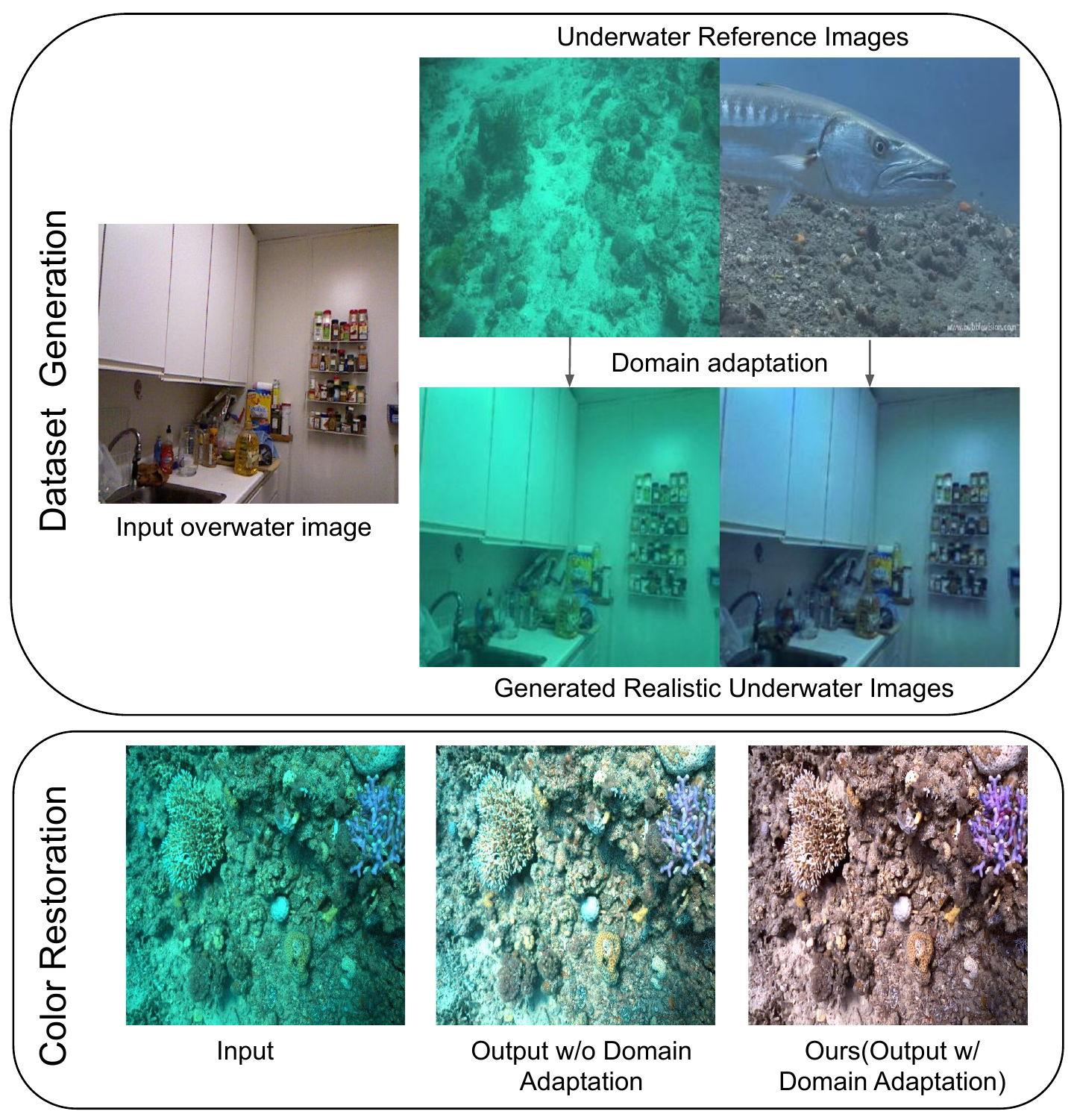}
    \caption{We present a novel underwater RGB dataset with diverse underwater conditions using multimodal domain adaptation. Our network when trained on domain adapted dataset produces better color restoration.  }
    \label{Main}
\end{figure}

\section{Introduction}
\label{sec:intro}

Images taken underwater have a blue or green hue due to the fact that the red wavelength gets absorbed in the water easily. Also, underwater images suffer from degradation due to the scattering of light by particles present in the water. Due to the aforementioned reasons, the visibility of the scene is quite less.  
Underwater color restoration aims to reconstruct the true color from images taken in underwater setting. This is essential for downstream tasks such as object recognition and semantic segmentation which have potential applications in monitoring sea life and marine research. Traditional methods~\cite{gdcp,blurriness,minimum_info} focus on inverting the underwater image formation model by considering different underwater and natural image priors. These priors fail to capture the complex underwater image formation model.

Another alternative to restore color are learning based methods ~\cite{Uplavikar_2019_CVPR_Workshops,UGAN,han2021cwr,Ucolor,unsupervised}. In recent years, Convolutional Neural Networks (CNNs) and Generative Adversarial Networks (GANs)  have shown great promise in a variety of low level computer vision tasks. 
The major challenge in color restoration of underwater images using learning based methods which are generally supervised is the absence of real-world datasets which contain the ground truth. Finding the underwater and above water versions of the same scene is extremely difficult. The simplified underwater image formation model used by existing learning based methods such as  \cite{li2019underwater,Uplavikar_2019_CVPR_Workshops,Ucolor} to create synthetic paired data is given by

\begin{equation}
I_{c}(x)=J_{c}(x) t_{c}(x)+A_{c}\left(1-t_{c}(x)\right)
\end{equation}

where $J_{c}$ is the clean image, $I_{c}$ is the generated synthetic underwater image, $t_{c}$ is transmission map which reflects the fraction of the light reaching the camera after reflecting from point $x$ in the scene, $A_{c}$ is the ambient light in the scene. $t_{c}$ can be expressed by:
\begin{equation}
t_{c}(x)=e^{-\beta_{c} d(x)}
\end{equation}
where $\beta_{c}$ is the wide-band attenuation coefficient and d(x) is the distance between the camera and the scene.
This model is similar to the haze formation model, except that the underwater model considers a different wide-band  attenuation  coefficient for each  color channel. The performances of data driven learning-based methods is highly dependent on the quality of the training data used by them. Models trained on synthetic datasets created using this equation do not perform well when tested on real underwater datasets. This can be attributed to the fact that synthetic underwater datasets fails to capture the wide array of underwater conditions present in real life underwater conditions.

Recently  Akkaynak \textit{et al}.~\cite{akkaynak2019sea}  showed  that  using  this  variant of haze formation model for creating synthetic underwater data introduces significant errors and formulated a revised underwater image formation model~\cite{akkaynak2018revised}. The revised model is dependant on reflectance, object range, beam attenuation coefficient which are hard to model and hence it is difficult to generate synthetic data using it.

\begin{figure*}[!t]
    \centering
    \includegraphics[width=\textwidth]{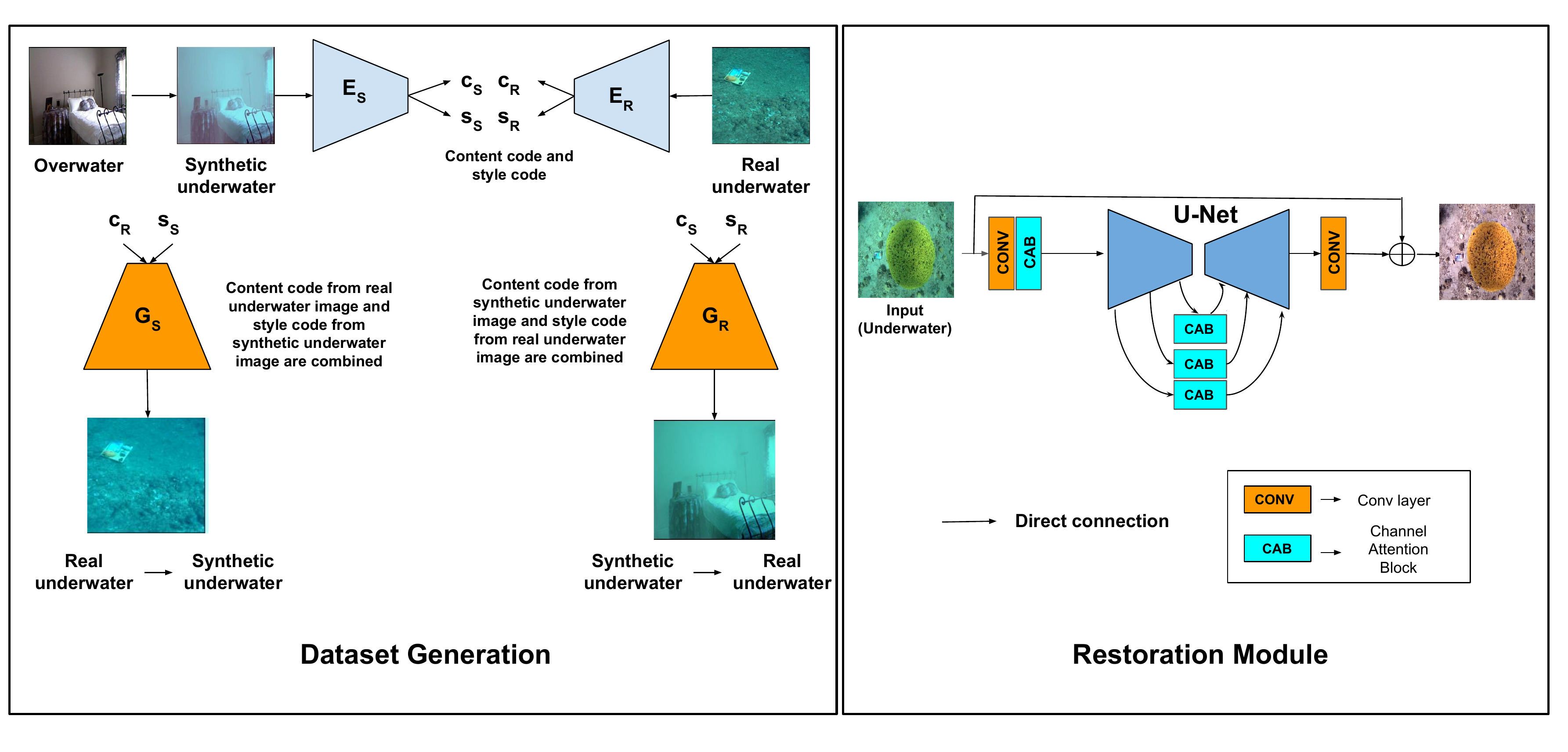}
    \caption{Overview of our proposed approach. \textbf{Dataset Generation}: First, we generate synthetic underwater image from aerial RGBD data using UWCNN \cite{li2019underwater}. Next, we extract content code from the synthetic image. To get diverse underwater images, we extract style code from multiple underwater images. Using content code and multiple style codes, our framework generates multiple realistic underwater images using multimodal domain adaptation \cite{huang2018multimodal}. \textbf{Color Restoration}: We use a residual U-net \cite{UNet} architecture with channel attention blocks for the restoration task.}
    \label{Network}
\end{figure*}

We attempt to address these challenges by using a multimodal unsupervised image-to-image translation framework ~\cite{huang2018multimodal}. 
To reduce the discrepancy between domains, our method first employs the
multimodal image translation network to translate images
from one domain to another.  To address the challenge of underwater image
distribution diversity, given a single image in the synthetic underwater domain,  our framework can produce many images corresponding to various styles of underwater images. Using this domain adaptation framework, we generate a more realistic underwater dataset. To the best of our knowledge, ours is the first method which applies multi-modal domain adaptation technique to generate a underwater image dataset which is not only diverse but is able to model real underwater conditions closely. To show the effectiveness of our novel underwater dataset, we train a simple but effective residual U-Net \cite{UNet} based architecture with channel attention blocks for the restoration task. Additionally, we observe that the performance of existing color restoration methods increases when trained our our domain adapted dataset. 
In summary, the main contributions of our paper are:
\begin{itemize}
    \item We present a novel domain adapted underwater RGB dataset and the corresponding paired ground truth. This dataset covers various different water conditions for better generalizability.
    \item We use a simple and effective residual U-Net architecture with channel attention blocks, which is inspired by \cite{Zamir2021MPRNet}, for performing underwater restoration on our domain adapted dataset.
    \item We conduct extensive experiments on our domain adapted dataset and show that our proposed method performs better than existing state of the art methods on color restoration tasks. 
\end{itemize}

\section{Related Work}

In this section, we briefly discuss the underwater color restoration methods which are related to our work. Depending upon the type of method used, underwater color restoration methods can be mainly categorized into either prior based methods, model-free methods or learning based methods. 

\subsection{Prior based methods}

Image prior based color restoration methods attempt to estimate the parameters of underwater imaging models based on statistics of clear abovewater images. Some of the priors employed by these methods include general dark channel prior (DCP) ~\cite{gdcp} which applied a general variant of the DCP to the underwater restoration problem. To estimate the transmission map of underwater images, Galdran \textit{et al.} \cite{GALDRAN2015132} proposed a variant of the DCP based on red channel information. Based on a minimum information loss and histogram distribution prior, Li et al. \cite{hdcp}, proposed an color restoration algorithm. Peng \textit{et al.} ~\cite{blurriness} proposed a depth estimation method for underwater scenes based on image blurriness and light absorption. Scene depth can be used to restore the degraded underwater image. SeaThru \cite{akkaynak2019sea} is a physics based algorithm which takes RGBD as input and estimates the backscatter and the illumination of the scene using priors. It then uses the revised underwater image formation model \cite{akkaynak2018revised} to recover the true radiance of the scene. The priors assumed by these methods are not valid in all underwater scenarios due to which their performance is limited. 

\subsection{Model-free methods}

Earlier attempts to restore color in underwater images involved adjusting the values of pixels to restore the color. They do not rely on any priors while restoring the image. Earlier solutions relied on histogram equalisation techniques \cite{histoeq} for restoration. Fusion based white balancing proposed by Ancuti \textit{et al.} ~\cite{8058463} proposed a two step strategy that involved white balancing and fusion to produce the color restored version. Fu \textit{et al.} \cite{7025927} proposed a retinex based approach for enhancing the underwater images.  To increase underwater image quality, Ghani et al. \cite{contrast} proposed a method to integrate global and local contrast stretching. Using these methods may result in distortion in colors and artifacts when illumination conditions are complex.



\subsection{Learning based methods}

There are several attempts to create a paired dataset which contains the underwater scene and the corresponding abovewater scene. Li \textit{et al}. ~\cite{li2019underwater} proposed to create a synthetic dataset using an underwater image formation model and created ten different types of underwater scenes based on the ten Jerlov water types. In~\cite{UGAN}, Fabbri \textit{et al}. proposed U-GAN approach which tries to translate an image from above water domain to underwater domain without predefined image pairs. It uses a CycleGAN based model to translate images from above water domain to under water domain. It then uses a CNN based model to restore images. Recently, Li \textit{et al}. ~\cite{li2019underwater} collected a paired underwater image dataset UIEBD using survey methods for training deep networks and proposed a gated fusion network to enhance underwater images. To collect this paired dataset, $12$ candidate ground-truth images for each underwater image were created using existing underwater image restoration methods. The authors used subjective comparisons among volunteers to select the best candidate image. Han \textit{et al}. \cite{han2021cwr} used physical apparatus and the camera response curve to measure the attenuation coefficients in water and used the old underwater image formation model to estimate the radiance of the scene. This dataset is limited in variety and has information about only one Jerlov water type. 

Other underwater color restoration methods use architectures and loss functions which are tailored to underwater domain. Upvaikar \textit{et al}. \cite{Uplavikar_2019_CVPR_Workshops} used a nuisance classifier to classify the type of underwater type in addition to a CNN based encoder-decoder architecture to restore images. Sharma \textit{et al.} \cite{sharma2021wavelength} used an attentive skip mechanism to refine multi scale features.  Li et al. \cite{Ucolor} used an attention mechanism to integrate features from different color spaces to produce a restored version of the underwater image. However, due to the domain gap between synthetic and real data, these methods which are trained on synthetic underwater images do not perform well when tested with real underwater images.

\begin{figure*}[!t]
    \includegraphics[width=\textwidth]{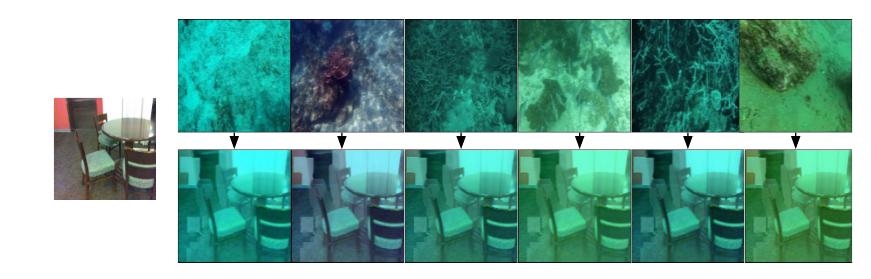}
    \caption{For each RGB-D input, we generate various UW images using the style of reference real UW images.}
    \label{Dataset}
\end{figure*}

\section{Proposed method}

In this section, we describe the details of our multimodal domain adaptation framework for underwater color restoration. First, we provide details of the multimodal image translation module which is used to create a more realistic dataset. Then we describe restoration module which is used to obtain the restored image. Finally, we give the loss functions that are applied to train the proposed networks. An overview of our proposed approach is shown in Fig. \ref{Network}.

\subsection{Multimodal Domain Adaptation}

Due to the domain gap, underwater restoration models trained on synthetic datasets are ineffective on real images. Domain adaptation techniques  helps learning based methods which are trained on a source domain to be tested on a different target domain without requiring any training annotations in the target domain. We perform domain adaptation between synthetic underwater images and a real underwater images to create our domain adapted dataset. Directly learning an image to image translation framework between real underwater images and above water images is not possible due to the huge gap between the two domains. Such a setup is susceptible to learning spurious correlations and does not yield good results. 

Given a set of synthetic underwater images $I_{S}= \{I_{i}\}_{i=1}^{N_{suw}}$ and a set of real underwater images $I_{R}=\{I_{j}\}_{j=1}^{N_{ruw}}$ where $N_{suw}$ and ${N_{ruw}}$ denote the number of images in the synthetic underwater and the real underwater domain respectively, we aim to learn a domain adaptation model that can produce a diverse set of underwater images that resembles the real underwater conditions.

We assume that an image can be decomposed into a content code which is domain invariant and a style code which captures the properties specific to a domain.  Let $E_{S}$ and $E_{R}$ be the encoders of the synthetic underwater and real underwater domain respectively. An encoder decomposes an image into a content code $c$ and a style code $s$.  Let $G_{S}$ and $G_{R}$ be the decoders of the synthetic underwater and real underwater domain respectively. Given a content code $c$ and a style code $s$, the decoder $G$ outputs the corresponding image. The decoder uses adaptive instance normalisation layers (AdaIN) \cite{huang2017adain} to combine the content and the style and generate an image. To ensure that the \textbf{synthetic underwater  $\rightarrow$ real underwater images} obtained upon translation are similar to the real underwater images and vice versa, we make use of discriminators $D_{R}$ and $D_{S}$ and use an adversarial setup to train our framework. The architecture of the encoder, generator and the discriminator follows that from MUNIT \cite{huang2018multimodal}.

\noindent
\textbf{Notation:} We use the following notation while defining our loss functions. Let $I_{suw} \in I_{S}$ and $I_{ruw} \in I_{R}$. To translate a synthetic underwater image $I_{suw}$ to the real underwater domain, we first generate the content code $c_{S}$ using $E_{S}$. We then generate the $s_{R}$ using $E_{R}$ of a real underwater image $I_{ruw}$. To translate the image, we use $G_{R}$ to recombine the content code $c_{S}$ and the style code $s_{R}$ to get $I_{S\rightarrow R}$.  

\begin{align*}
\begin{split}    
    c_{S},s_{S} = E_{S}(I_{suw}) \\ 
    c_{R},s_{R} = E_{R}(I_{ruw}) \\
    I_{suw\rightarrow ruw} = G_{R}(c_{S},s_{R}) \\
\end{split}
\end{align*}

\begin{center}
\end{center}

\begin{center}
\end{center}
\begin{figure*}[!t]
    \begin{tabularx}{\textwidth }{ 
  >{\centering\arraybackslash}X >{\centering\arraybackslash}X >{\centering\arraybackslash}X >{\centering\arraybackslash}X >{\centering\arraybackslash}X >{\centering\arraybackslash}X 
   }
    Input & DWN~\cite{sharma2021wavelength}  & FUnIE-GAN~\cite{islam2020fast}  & WaterNet~\cite{li2019underwater} & Ucolor~\cite{Ucolor} & Ours \\ 
    \end{tabularx}
    \centering
    \includegraphics[width=\textwidth]{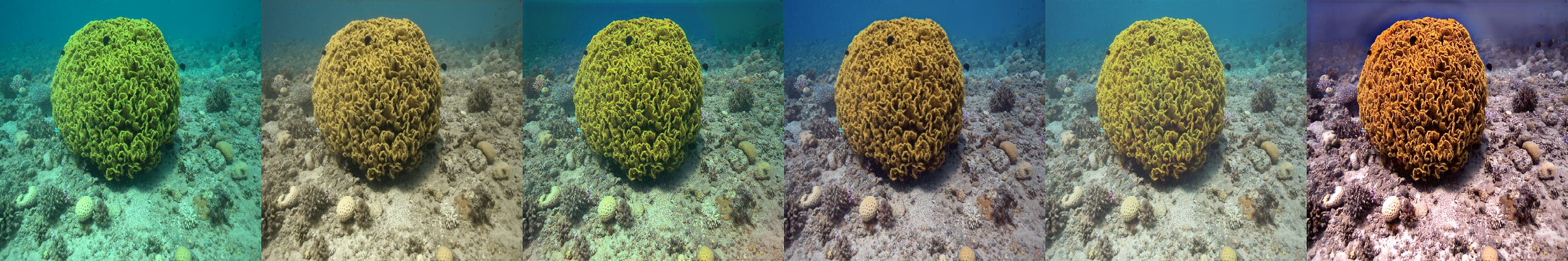}
        \begin{tabularx}{\textwidth }{ 
  >{\centering\arraybackslash}X >{\centering\arraybackslash}X >{\centering\arraybackslash}X >{\centering\arraybackslash}X >{\centering\arraybackslash}X >{\centering\arraybackslash}X 
   }
    Input & DWN~\cite{sharma2021wavelength}  & FUnIE-GAN~\cite{islam2020fast}  & WaterNet~\cite{li2019underwater} & Ucolor~\cite{Ucolor} & Ours \\ 
    \end{tabularx}
    \includegraphics[width=\textwidth]{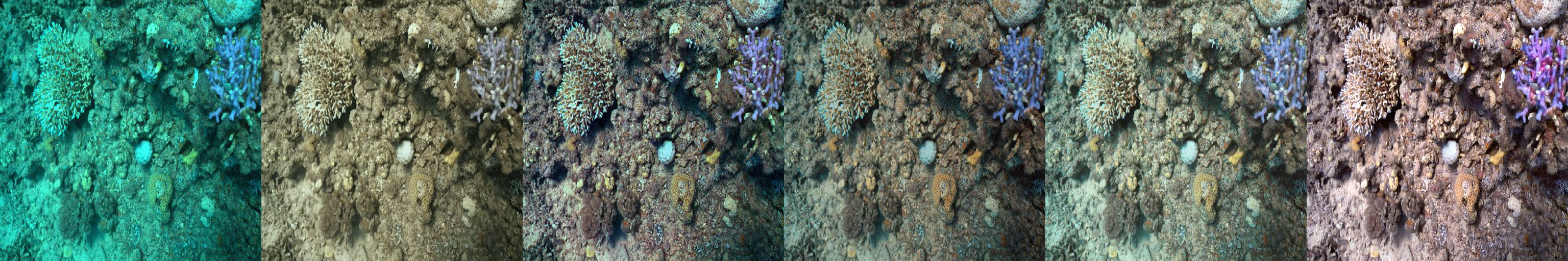}
        \begin{tabularx}{\textwidth }{ 
  >{\centering\arraybackslash}X >{\centering\arraybackslash}X >{\centering\arraybackslash}X >{\centering\arraybackslash}X >{\centering\arraybackslash}X >{\centering\arraybackslash}X 
   }
    Input & DWN~\cite{sharma2021wavelength}  & FUnIE-GAN~\cite{islam2020fast}  & WaterNet~\cite{li2019underwater} & Ucolor~\cite{Ucolor} & Ours \\ 
    \end{tabularx}
    \includegraphics[width=\textwidth]{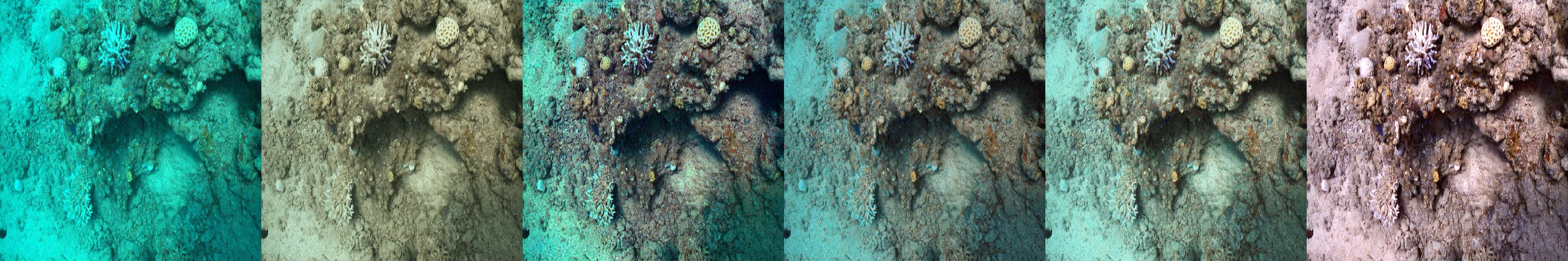}
            \begin{tabularx}{\textwidth }{ 
  >{\centering\arraybackslash}X >{\centering\arraybackslash}X >{\centering\arraybackslash}X >{\centering\arraybackslash}X >{\centering\arraybackslash}X >{\centering\arraybackslash}X 
   }
    \end{tabularx}
    \caption{Comparison of our method with recent state of the art methods on the SeaThru \cite{akkaynak2019sea} dataset. Our method is able to handle underwater color attenuation well and can restore it properly.}
    \label{Comparisons}
\end{figure*}
\vspace{-0.5cm}
\subsubsection{Loss functions used for Multimodal domain adaptation}

To train our multimodal domain adaptation framework, we use the following loss functions:

\noindent
\textbf{Style and Content Reconstruction loss:} We make use of cyclic reconstruction losses for the latent codes to ensure that the encoder and decoder for each of the domains are inverses of one another. The loss functions for style and content reconstruction are:

\begin{align}
\begin{split}
        \mathcal{L}^{R}_{\text{content}} =  \| E_{R}^{c}(G_{R}(c_{S},s_{R})) - c_{S} \|_{1} \\
        \mathcal{L}^{R}_{\text{style}} =  \| E_{R}^{s}(G_{R}(c_{S},s_{R})) - s_{R} \|_{1}
\end{split}
\end{align}

\noindent
\textbf{Identity mapping loss for images:} Using identity mapping losses for images also constraints the encoder and decoder for each of the domains to be inverses of one another. The loss functions used for identity mapping loss are:

\begin{align}
\begin{split}
        \mathcal{L}^{R}_{\text{img}} =  \| G_{R}(E_{R}^{c}(I_{ruw}),E_{R}^{s}(I_{ruw}))  - I_{ruw} \|_{1}
\end{split}
\end{align}

\noindent
\textbf{Adversarial Loss:} The adversarial loss reduces the discrepancy between the distribution of the real underwater images and the translated synthetic underwater to real underwater images. This loss is essential to ensure that the real underwater images and the translated synthetic are indistinguishable.
\begin{align}
\begin{split}
        \mathcal{L}^{R}_{\text{adv}} = log(1-D_{R}(G_{R}(c_{S},s_{R})) + log(D_{R}(I_{ruw})) 
\end{split}
\end{align}

\noindent
\textbf{Edge Loss:} While translating images from the synthetic to real domain, we want to preserve edges and the structure of the synthetic underwater image. To calculate the edge loss, we first apply laplacian of the gaussian to extract the edges and then enforce L1 loss between the translated image and the original image.

\begin{align}
\begin{split}
    \mathcal{L}^{R}_{\text{edge}} = \sqrt{(\nabla^2(\hat{I}_{suw})- \nabla^2(\hat{I}_{suw \rightarrow ruw}))^2+\varepsilon^2}
\end{split}
\end{align}

$\nabla^2$ denotes the Laplacian operator and $\hat{I}_{suw}$, $\hat{I}_{suw \rightarrow ruw}$ are the Gaussian smoothed form of ${I_{suw}}$, ${I_{suw \rightarrow ruw}}$ respectively . We set the value of $\varepsilon$ to $10^{-4}$ in our experiments.

\noindent
\textbf{SSIM Loss:} To enforce the translation module to use the input image as a reference while translating from synthetic to real, we calculate the SSIM score between the input image and the translated version of it. 

\begin{align}
\begin{split}
    \mathcal{L}^{R}_{\text{SSIM}} = 1 - SSIM(I_{suw \rightarrow ruw},I_{suw})
\end{split}
\end{align}

\noindent
We define the other loss functions such as $\mathcal{L}^{S}_{\text{content}}$, $\mathcal{L}^{S}_{\text{style}}$, $\mathcal{L}^{S}_{\text{adv}}$, $\mathcal{L}^{S}_{\text{SSIM}}$,$\mathcal{L}^{S}_{\text{edge}}$ for the translated images of the real underwater domain to the synthetic underwater domain symmetrically.

\begin{figure*}[!t]
    \begin{tabularx}{\textwidth }{ 
  >{\centering\arraybackslash}X >{\centering\arraybackslash}X >{\centering\arraybackslash}X >{\centering\arraybackslash}X >{\centering\arraybackslash}X >{\centering\arraybackslash}X 
   }
    Input & DWN~\cite{sharma2021wavelength}  & FUnIE-GAN~\cite{islam2020fast}  & WaterNet~\cite{li2019underwater} & Ucolor~\cite{Ucolor} & Ours \\ 
    \end{tabularx}
    \centering
    \includegraphics[width=\textwidth]{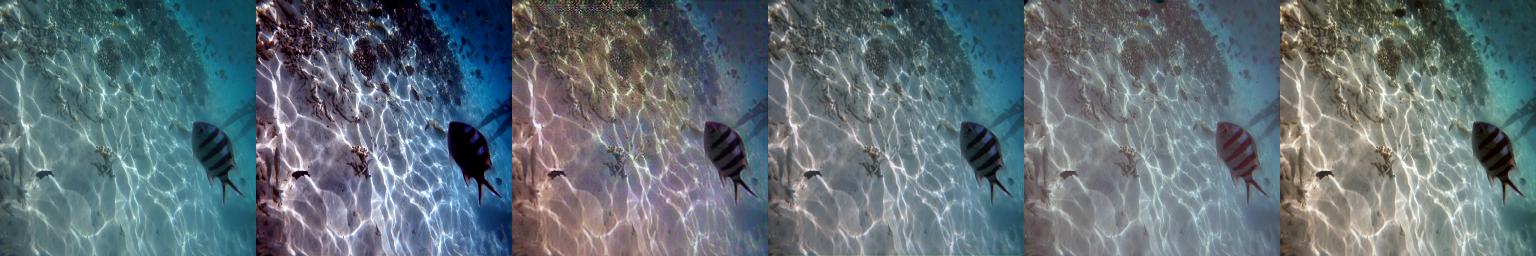}
        \begin{tabularx}{\textwidth }{ 
  >{\centering\arraybackslash}X >{\centering\arraybackslash}X >{\centering\arraybackslash}X >{\centering\arraybackslash}X >{\centering\arraybackslash}X >{\centering\arraybackslash}X 
   }
    Input & DWN~\cite{sharma2021wavelength}  & FUnIE-GAN~\cite{islam2020fast}  & WaterNet~\cite{li2019underwater} & Ucolor~\cite{Ucolor} & Ours \\ 
    \end{tabularx}
    \includegraphics[width=\textwidth]{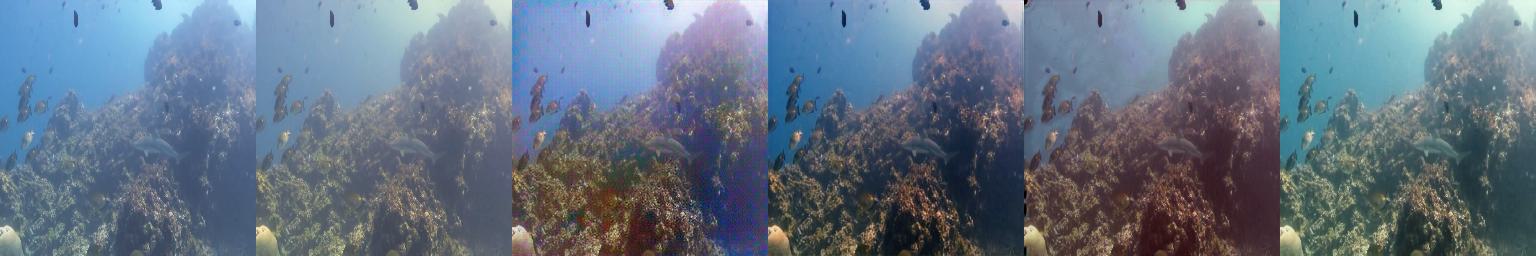}
        \begin{tabularx}{\textwidth }{ 
  >{\centering\arraybackslash}X >{\centering\arraybackslash}X >{\centering\arraybackslash}X >{\centering\arraybackslash}X >{\centering\arraybackslash}X >{\centering\arraybackslash}X 
   }
    Input & DWN~\cite{sharma2021wavelength}  & FUnIE-GAN~\cite{islam2020fast}  & WaterNet~\cite{li2019underwater} & Ucolor~\cite{Ucolor} & Ours \\ 
    \end{tabularx}
    \includegraphics[width=\textwidth]{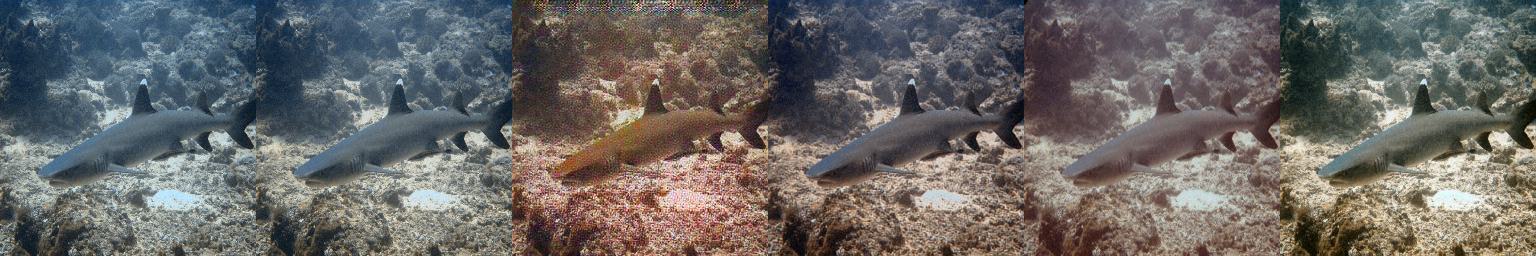}
            \begin{tabularx}{\textwidth }{ 
  >{\centering\arraybackslash}X >{\centering\arraybackslash}X >{\centering\arraybackslash}X >{\centering\arraybackslash}X >{\centering\arraybackslash}X >{\centering\arraybackslash}X 
   }
    \end{tabularx}
    \caption{Visual Comparisons on the real UIEBD \cite{li2019underwater} dataset}
    \label{comparisons_uiebd}
\end{figure*}

\noindent
\textbf{Overall Loss Function:} Thus, the overall loss function is: 

\begin{align}
\begin{split}
 \min_{E_{R}, E_{S}, G_{R}, G_{S}} & \max_{D_{R}, D_{S}}  \mathcal{L}\left(E_{R}, E_{S}, G_{R}, G_{S}, D_{R}, D_{S}\right) = \\ & \lambda_{content} (\mathcal{L}^{S}_{\text{content}} + \mathcal{L}^{R}_{\text{content}}) + \lambda_{style} (\mathcal{L}^{S}_{\text{style}} + \mathcal{L}^{R}_{\text{style}}) \\ & +  \lambda_{img} (\mathcal{L}^{S}_{\text{img}} + \mathcal{L}^{R}_{\text{img}})  +  \lambda_{adv} (\mathcal{L}^{S}_{\text{adv}} + \mathcal{L}^{R}_{\text{adv}}) \\  &+ \lambda_{edge} (\mathcal{L}^{S}_{\text{edge}} +
\mathcal{L}^{R}_{\text{edge}}) +\lambda_{ssim} (\mathcal{L}^{S}_{\text{SSIM}} + \mathcal{L}^{R}_{\text{SSIM}})
\end{split}
\end{align}

\noindent

where $\lambda_{content}$, $\lambda_{style}$, $\lambda_{img}$,$\lambda_{adv}$,  $\lambda_{ssim}$ and $\lambda_{edge}$ control the relative importance between the losses. Using the above setup, we try to learn a mapping between synthetic underwater domain and real underwater domain.

\subsection{Restoration framework}

We use an encoder-decoder framework which is based on the U-Net architecture \cite{UNet}. To exploit the inter-channel relationship of features, our framework consists of channel attention blocks (CAB) \cite{10.1007/978-3-030-01234-2_1}. We use CABs at every scale as well as at the skip connections of the U-Net. These blocks are capable of extracting multi-scale contextualized features \cite{Zamir2021MPRNet} quite well. It has been observed that using transposed convolution operation in the decoder part introduces checkerboard artifacts in the restored image. Therefore,  to increase the spatial resolution of features in the decoder, we use bilinear upsampling followed by a convolution layer. Finally, instead of directly predicting the restored image, our proposed restoration framework predicts a residual image $I_R$ to which the degraded input image $I_X$ is added to obtain: $I_{Y}$ = $I_X$ + $I_R$. where $I_{G}$ denotes the ground truth image.

\subsubsection{Loss Functions used for Restoration}

We use a combination of Charbonnier's loss \cite{charbo}, perceptual loss \cite{Johnson2016Perceptual} and edge loss to optimise our restoration network. Charbonnier's loss is an extension of the standard L1 loss which adds robustness to the optimization process. Perceptual loss encourages the network to produce natural looking images. Edge loss helps in preserving the high frequency details of the image. Thus, an optimal combination of these losses helps in restoring underwater images. 

\begin{multline}
    \emph{L} = \sqrt{(I_{G}- I_{Y})^2+\varepsilon^2} + \lambda_{1} L_{perceptual}(I_{G},I_{Y})+\\ \lambda_{2} \sqrt{(\nabla^2(\hat{I_{G}})- \nabla^2(\hat{I_{Y}}))^2+\varepsilon^2},
    \label{eq:loss1}
\end{multline}

where $I_{G}$ denotes the ground truth image, $I_{Y}$ denotes the output of our model,  $\nabla^2$ denotes the Laplacian operator and $\hat{I_{Y}}$, $\hat{I_{G}}$ are the Gaussian smoothed form of ${I_{Y}}$, ${I_{G}}$ respectively . We set the value of $\varepsilon$ to $10^{-4}$ in our experiments. Here $\lambda_{1}$ and $\lambda_{2}$ control the relative importance of the three losses used.

\begin{table}
    \centering
    \begin{tabular}{|c | c  |} 
     \hline
     Datasets & \# of images \\ [0.5ex] 
     \hline\hline
     SeaThru \cite{akkaynak2019sea} & 1157  \\ 
     \hline
     HICRD \cite{han2021cwr}  & 2003 \\
     \hline
     RUIE \cite{ruie}  & 2181 \\
     \hline
     UIEBD \cite{uieqb} & 890  \\
     \hline
     SQUID \cite{berman2020underwater}  & 57 \\
     \hline
     \hline
     Total & 6288 \\ [1ex]
     \hline
    \end{tabular}
    \caption{Details regarding the real underwater datasets used in our work. No single real-world dataset features a large number of diverse underwater conditions. Therefore, to increase diversity, we use many sources.}
    \label{dataset_images}
\end{table}

\begin{table}[!h]
  \centering
  \begin{tabular}{|c | c | c |} 
   \hline
   Methods & UCIQE $\uparrow$ & UIQM $\uparrow$ \\ [0.5ex] 
   \hline\hline
   Input & 0.4663 & 2.0196  \\ 
   \hline
   DWN & 0.4582 &  4.9045 \\
   \hline
   FUnIE-GAN & 0.503 & 4.8331  \\
   \hline
   WaterNet & 0.5331 & 5.079 \\
   \hline
   Ucolor & 0.5218 & 4.7468 \\
   \hline
   Ours(simple U-Net) & 0.5363 & 5.7806 \\
   \hline
   Ours(w/o CAL) & 0.5419 & 5.7851 \\ 
   \hline
   Ours & \textbf{0.5642} & \textbf{6.0271} \\ [1ex]
   \hline
  \end{tabular}
\caption{UnderWater quantitative metrics (higher is better) for various methods when tested on SeaThru dataset \cite{akkaynak2019sea}.}
\label{Quant}
\end{table}

\vspace{-1cm}
\section{Experimental Details}

In this section, we first present the details on how we create our domain adapted dataset. Secondly, we provide details regarding training of our framework. Then, we provide quantitative and qualitative evaluations of our methods on real underwater images. Finally, we perform ablation studies to verify the effectiveness of our approach.

\begin{figure*}[!t]
    \centering
    \includegraphics[width=\textwidth]{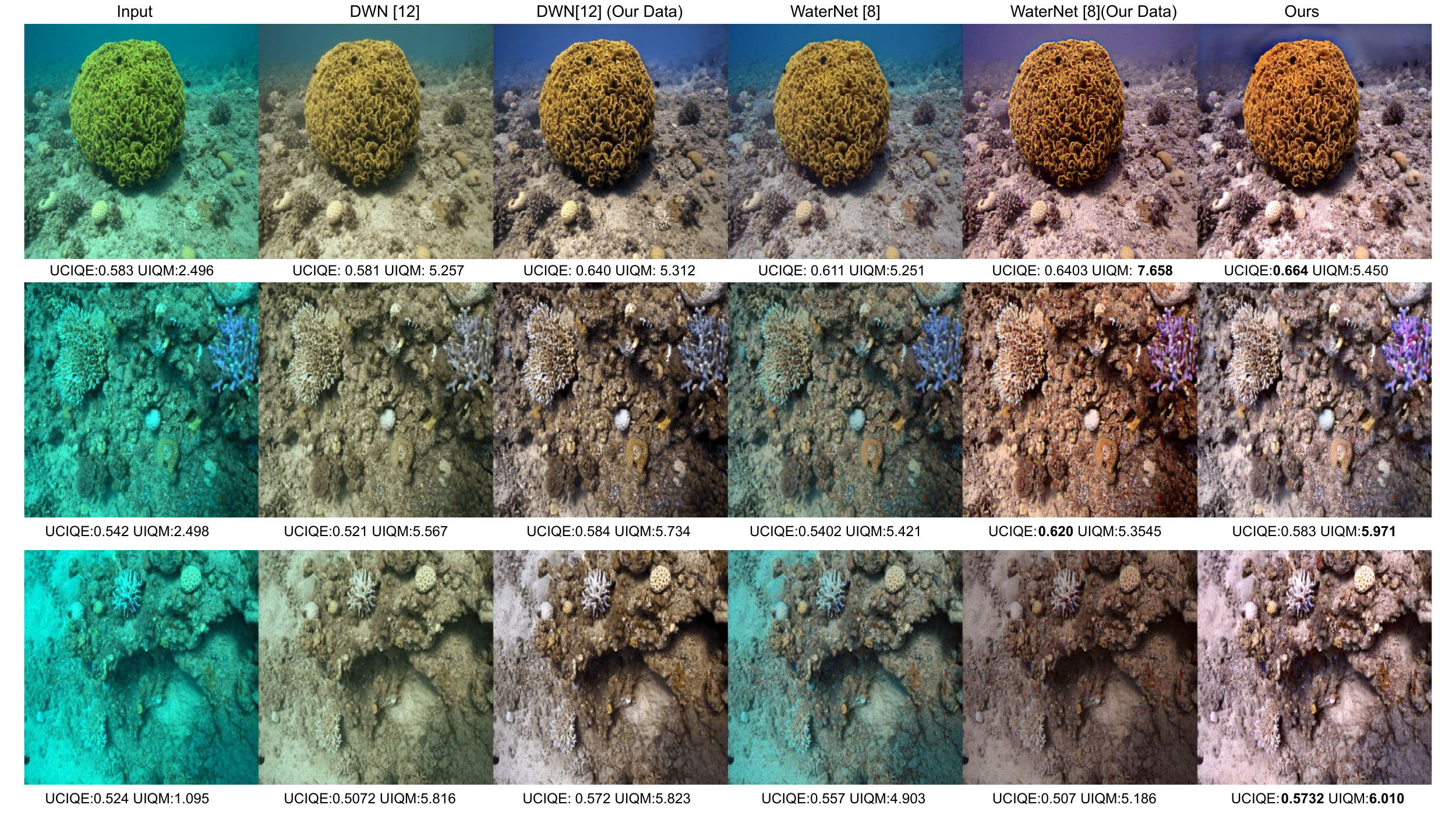}
    \caption{We compare the performance of various methods when trained on our domain adapted dataset versus the original dataset that the methods were trained on. 
trained on }
    \label{comparison_our_ds}
\end{figure*}

\subsection{Creating domain adapted dataset}

\begin{figure}[htp]
    \centering
    \includegraphics[width=\columnwidth]{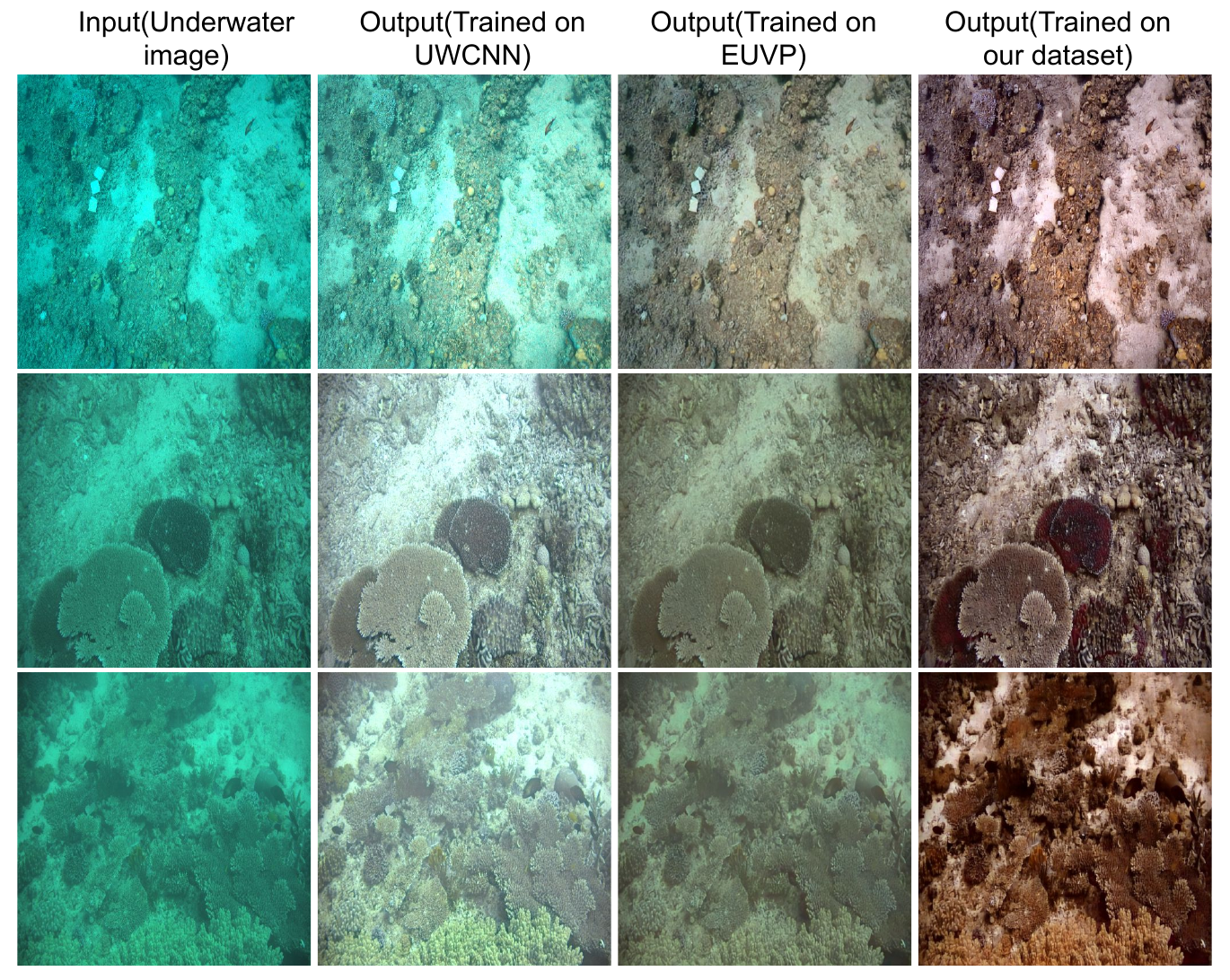}
    \caption{We train our network on other synthetic datasets such as UWCNN \cite{li2019underwater} and EUVP \cite{UGAN}. Network trained on our domain adapted dataset produces better results.}
    \label{Ablation}
\end{figure}
To perform domain adaptation, we need synthetic underwater data and real underwater data. We use $1449$ RGBD images from NYU-V2 dataset \cite{Silberman:ECCV12} to create synthetic underwater images according to the procedure given by \cite{li2019underwater}. We use the Type I and Type-III variants in our synthetic data. To create the real underwater dataset, we collate an extensive set of images from many different sources which cover a diverse range of underwater conditions. This allows our model to generalise well to unseen water types. Further details, regarding the sources and the number of images can be found in Table \ref{dataset_images}. We train the multimodal domain adaptation framework for this \textbf{synthetic underwater dataset $\leftrightarrow$ real underwater dataset} without using any paired information.

To generate our domain adapted dataset, we extract the content code from a synthetic underwater image and extract the style code from a randomly sampled real underwater image and perform domain adaptation using them, see Fig. \ref{Dataset}. More examples from our domain adapted dataset can be found in the supplementary material. For each synthetic underwater image, we randomly sample $6$ real underwater images from multiple real underwater images and extract the style code for each of these $6$ images. Using these styles codes and the content codes extracted from the synthetic underwater dataset, we create our domain adapted dataset.  Therefore, in total we have $1449\times6=8694$ images in our domain adapted dataset and their corresponding ground truth. Using this domain adapted dataset and the corresponding ground truth, we train our restoration framework on it. We plan to make this domain adapted dataset available for academic purposes and for further research.

\subsection{Training details}

The experiments for multimodal domain adaptation and the restoration framework were performed on Nvidia RTX 2080Ti using PyTorch framework.

\subsubsection{Multimodal domain adaptation}

To train our framework, we use images of size $128x128$. In all experiments, we use a batch size of 4 and train the framework for $10^5$ steps until all the loss functions stabilise. The weight parameters $\lambda_{content}$, $\lambda_{style}$, $\lambda_{img}$, $\lambda_{adv}$, $\lambda_{ssim}$, $\lambda_{edge}$ are set to $1$, $1$, $1$, $5$, $3$ and $50$ respectively. 

\subsubsection{Restoration}

For our restoration framework, we use patches of size $256 \times 256$.  Horizontal and vertical flips are randomly applied as forms of data augmentation. We train the restoration framework for $75$ epochs. The weight parameters $\lambda_{1}$ and $\lambda_{2}$ are set to 0.5 and 0.5 respectively. We use the Adam optimizer with an initial learning rate of a $ 3\times10^{-4}$ which is halved after every $10$ epochs.

\begin{table}
  \centering
  \begin{tabular}{|c | c | c |} 
   \hline
   Methods & PSNR $\uparrow$ & SSIM $\uparrow$ \\ [0.5ex] 
   \hline\hline
   Input & 16.91 & 0.761 \\ 
   \hline
   DWN & 21.52 &  0.801 \\
   \hline
   FUnIE-GAN & 18.86 & 0.782  \\
   \hline
   WaterNet & 19.24 & 0.832 \\
   \hline
   Ucolor & 21.63 & 0.838 \\
   \hline
   Ours & \textbf{21.83} & \textbf{0.844} \\ [1ex]
   \hline
  \end{tabular}
\caption{Full-reference quantitative metrics for various methods when tested on the UIEBD Dataset\cite{li2019underwater}} 
\label{quant_full_reference}
\end{table}

\subsection{Comparison with state-of-the-art methods}

We compare our method with existing state of the art deep methods \cite{sharma2021wavelength,islam2020fast,li2019underwater,Ucolor}. We test our method on the SeaThru \cite{akkaynak2019sea} dataset and UIEBD \cite{li2019underwater} dataset. As reported in Fig. \ref{Comparisons}, our method is able to remove the greenish/bluish hue from underwater images in the SeaThru dataset. Since ground truth data is not available for the SeaThru dataset, non reference underwater image restoration metrics like Underwater Color Image Quality Evaluation Metric (UCIQE) \cite{yang2015underwater} and 
Underwater Image Quality Measure (UIQM) \cite{panetta2015human} are calculated and reported in Table \ref{Quant}. 


Visual comparisons on the UIEBD dataset are reported in Fig. \ref{comparisons_uiebd}. Since, ground truth data is available for UIEBD, full reference image restoration metrics like PSNR, SSIM \cite{1284395} are calculated and reported in Table \ref{quant_full_reference}. We can see that our method performs better than the existing state of the art methods and produces more natural looking images. More color restoration comparisons against other underwater restoration methods are included in the supplementary material.

\subsection{Training other methods on our domain adapted dataset}

Our main contribution lies in the fact that we create a multimodal domain adapted dataset that reflects the underwater conditions closely. To verify the effectiveness of our dataset, we train other deep learning based methods such as \cite{li2019underwater},\cite{sharma2021wavelength} on our dataset. We found that when trained on our domain adapted dataset, these methods perform better (both qualitatively and quantitatively) on real images,  than when trained on their own synthetic dataset. The qualitative results of this experiment can been seen in Fig.\ref{comparison_our_ds} and quantitative results in Table \ref{metrics_our_ds}, where we have calculated non-reference underwater image metrics on the real SeaThru \cite{akkaynak2019sea} dataset. This experiment also validates our claim that methods which are trained on synthetic underwater datasets do not perform well when tested on real underwater datasets. However, when we train these algorithms on our domain adapted dataset, then the performance of these algorithms increase.

\subsection{Ablation study}
To demonstrate the efficacy of our dataset, we trained our network with UWCNN~\cite{li2020underwater} synthetic dataset and with a GAN generated dataset EUVP ~\cite{islam2020fast}. Restoration results by our network trained on the UWCNN dataset is shown in second column of Fig. \ref{Ablation}. We can clearly see some green tinge in the output. EUVP dataset is a GAN based dataset which has 11435 paired examples. The restoration results by our method trained on this dataset is shown in the third column. The results are slightly better than that of UWCNN. This can be attributed to the fact that the EUVP dataset was created using a GAN and can model the underwater conditions better.  Overall, our method trained on our domain adapted dataset produces the best restoration results which is seen in the fourth column. 

We also tried out different variations of our network; 1) Simple U-Net, where we removed the residual connection and the channel attention layer from the channel attention block (CAB contains two convolutional layers followed by channel attention), 2) w/o Channel Attention Layer (CAL). All our variations perform better than existing methods. The results are shown in Table \ref{Quant}. Additional ablation studies on the losses used for the domain adaptation can be found in the supplementary material.

\begin{table}
    \centering
    \begin{tabular}{|c | c | c |} 
     \hline
     Methods & UCIQE $\uparrow$ & UIQM $\uparrow$ \\ [0.5ex] 
     \hline\hline
     Input & 0.4663 & 2.0196  \\ 
     \hline
     DWN [12] & 0.4582 &  4.9045 \\
     \hline
     DWN (\textbf{Our Data})  & \textbf{0.5375} & \textbf{5.6910}  \\
     \hline
     WaterNet \cite{li2019underwater} & 0.5331 & 5.079 \\
     \hline
     WaterNet \cite{li2019underwater} (\textbf{Our Data}) & \textbf{0.5421} & \textbf{5.84} \\
     \hline
     Ours & \textbf{0.5642} & \textbf{6.0271} \\ [1ex]
     \hline
    \end{tabular}
    \caption{Underwater quantitative metrics (higher is better) for various methods when tested on real SeaThru dataset. \textbf{Our Data} indicates that 
    the particular algorithm was trained on our domain adapted dataset}
    \label{metrics_our_ds}
\end{table}

\section{Conclusions}

In this paper, we create a diverse underwater image dataset which reflects the real underwater conditions closely. To create this dataset, we use a multimodal domain adaptation framework to translate an image from the synthetic underwater domain to the real underwater domain. We proposed a simple residual U-net based architecture with channel attention modules for color restoration. We show that our method produces state of the art results qualitatively and also quantitatively on real underwater datasets. By performing extensive experiments on our domain adapted dataset, we show that our method increases the performance of existing underwater restoration algorithms. In the future, we plan to use the multimodal domain adaptation framework to solve other image restoration tasks such as low light restoration, reflection removal where obtaining the ground truth is not possible.  
\begin{acks}
We gratefully acknowledge funding from the Department of Science and Technology (DST/ICPS/IHDS/2018).
\end{acks}
\bibliographystyle{ACM-Reference-Format}
\bibliography{ICVGIP-Latex-Template}

\end{document}


\title{Supplementary Material for "Towards Realistic Underwater Dataset Generation and Color Restoration"}

\author{Neham Jain}
\affiliation{%
  \institution{IIT Madras}
  \city{Chennai}
  \state{Tamil Nadu}
  \country{India}
  \postcode{600036}
}
\email{nehamjain2002@gmail.com}
\author{Gopi Raju Matta}
\affiliation{%
  \institution{IIT Madras}
  \city{Chennai}
  \state{Tamil Nadu}
  \country{India}
  \postcode{600036}
}
\email{ee17d021@smail.iitm.ac.in}

\author{Kaushik Mitra}
\affiliation{%
  \institution{IIT Madras}
  \city{Chennai}
  \state{Tamil Nadu}
  \country{India}
  \postcode{600036}
}
\email{kmitra@ee.iitm.ac.in}

\renewcommand{\shortauthors}{}

\maketitle

\section{More studies for our paper}
We provide some more examples of our domain adapted dataset that we train our restoration model on. We provide additional results on real world underwater images. Additionally, we perform an ablation study on the SSIM loss and Edge loss used in the domain adaptation framework.  Also, we show some failure cases for our model. 

\subsection{Examples of our domain adapted dataset}
\begin{figure*}[!t]
    \begin{tabularx}{\textwidth }{ 
  >{\centering\arraybackslash}X >{\centering\arraybackslash}X >{\centering\arraybackslash}X >{\centering\arraybackslash}X
   }
    Clean & Synthetic Underwater & After domain adaptation & Style Image \\ 
    \end{tabularx}
    \centering
    \includegraphics[width=\textwidth]{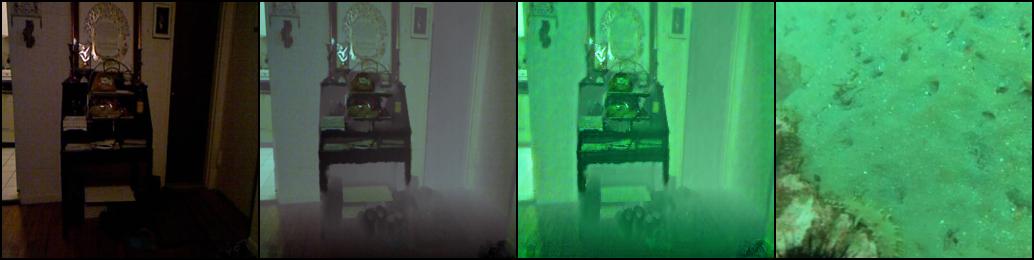}
    \includegraphics[width=\textwidth]{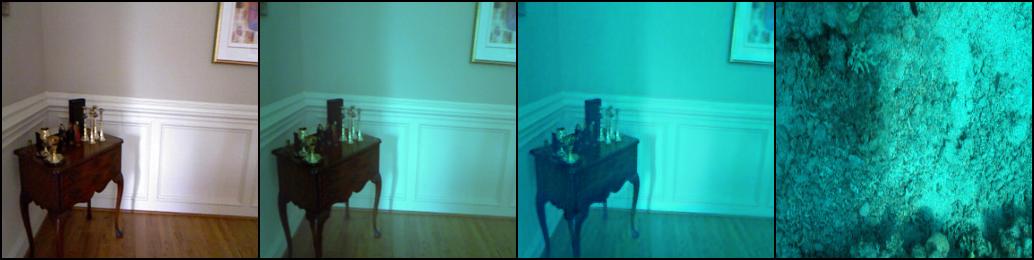}
    \includegraphics[width=\textwidth]{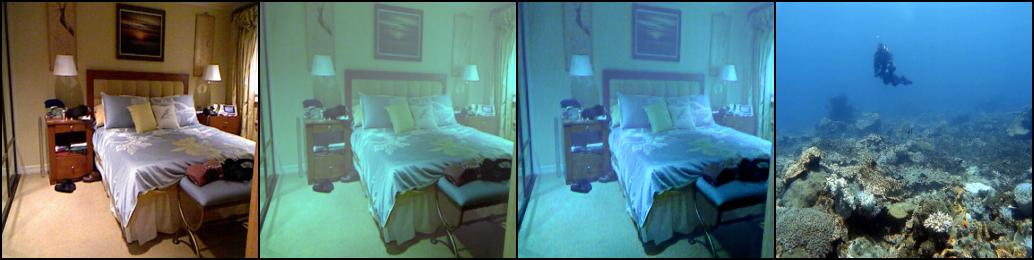}
    \includegraphics[width=\textwidth]{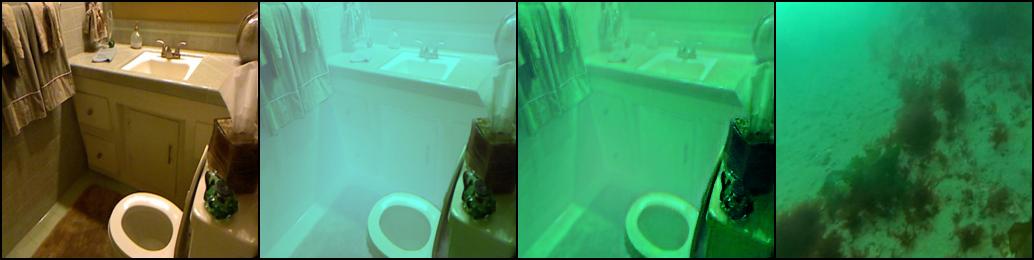}
    \includegraphics[width=\textwidth]{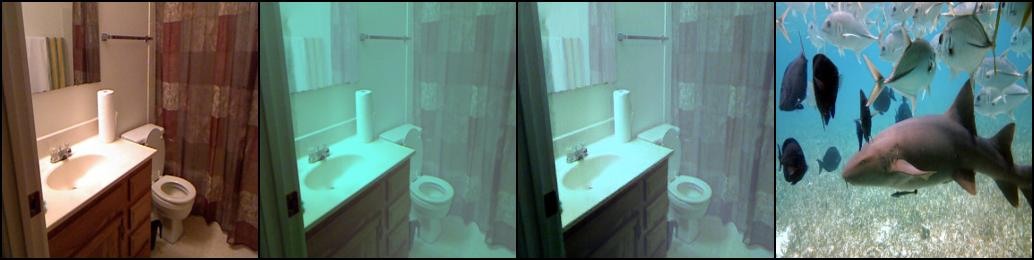}
    \caption{Some examples from our domain adapted dataset}
    \label{examples_dataset}
\end{figure*}

To create a more realistic underwater image, we combine the content code of a synthetic underwater image with the style code of a real underwater image. We provide some more examples for our domain adapted dataset in Fig. \ref{examples_dataset}

\subsection{Comparisons with state-of-the-art methods}

\begin{figure*}[!b]
    \begin{tabularx}{\textwidth }{ 
  >{\centering\arraybackslash}X >{\centering\arraybackslash}X >{\centering\arraybackslash}X >{\centering\arraybackslash}X >{\centering\arraybackslash}X >{\centering\arraybackslash}X 
   }
    Input & DWN~\cite{sharma2021wavelength}  & FUnIE-GAN~\cite{islam2020fast}  & WaterNet~\cite{li2019underwater} & Ucolor~\cite{Ucolor} & Ours \\ 
    \end{tabularx}
    \centering
    \includegraphics[width=\textwidth]{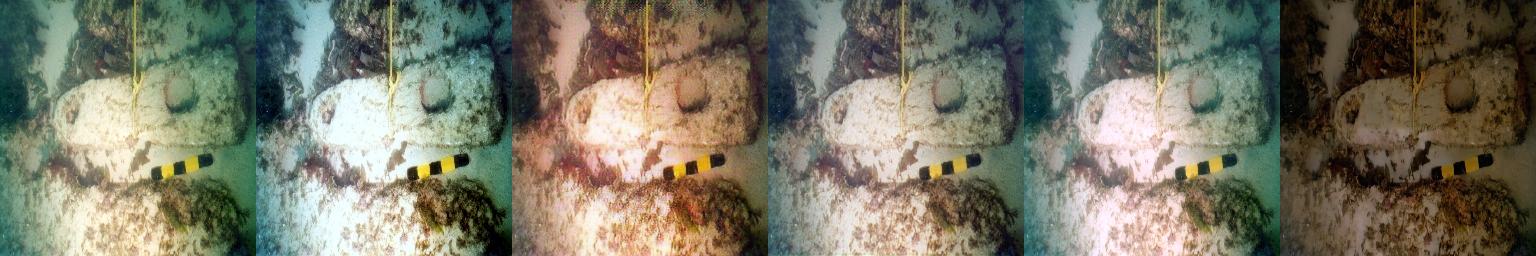}
        \begin{tabularx}{\textwidth }{ 
  >{\centering\arraybackslash}X >{\centering\arraybackslash}X >{\centering\arraybackslash}X >{\centering\arraybackslash}X >{\centering\arraybackslash}X >{\centering\arraybackslash}X 
   }
    Input & DWN~\cite{sharma2021wavelength}  & FUnIE-GAN~\cite{islam2020fast}  & WaterNet~\cite{li2019underwater} & Ucolor~\cite{Ucolor} & Ours \\ 
    \end{tabularx}
    \includegraphics[width=\textwidth]{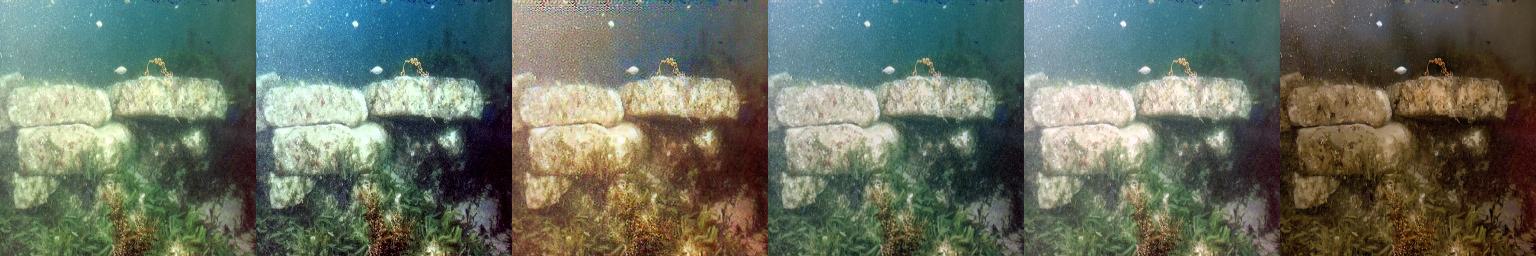}
        \begin{tabularx}{\textwidth }{ 
  >{\centering\arraybackslash}X >{\centering\arraybackslash}X >{\centering\arraybackslash}X >{\centering\arraybackslash}X >{\centering\arraybackslash}X >{\centering\arraybackslash}X 
   }
    Input & DWN~\cite{sharma2021wavelength}  & FUnIE-GAN~\cite{islam2020fast}  & WaterNet~\cite{li2019underwater} & Ucolor~\cite{Ucolor} & Ours \\ 
    \end{tabularx}
    \includegraphics[width=\textwidth]{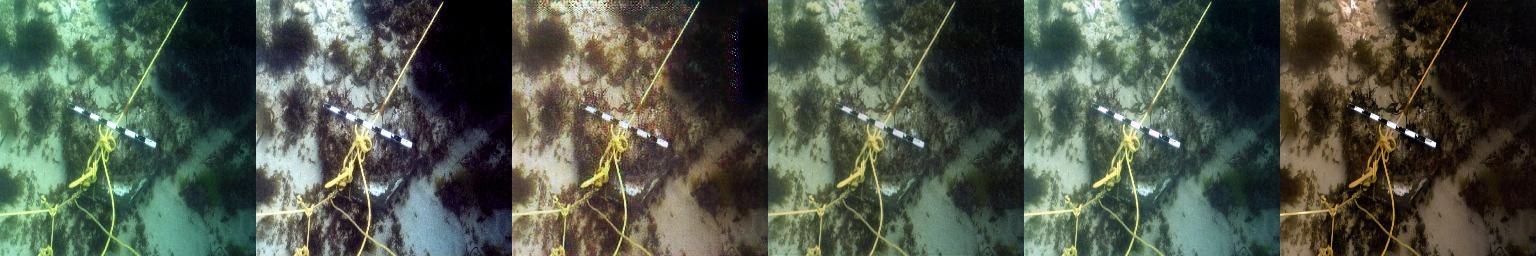}
            \begin{tabularx}{\textwidth }{ 
  >{\centering\arraybackslash}X >{\centering\arraybackslash}X >{\centering\arraybackslash}X >{\centering\arraybackslash}X >{\centering\arraybackslash}X >{\centering\arraybackslash}X 
   }
    \end{tabularx}
    \caption{Visual comparisons on the dataset collected by us during an expedition}
    \label{Comparisons}
\end{figure*}

\begin{figure*}[!t]
    \begin{tabularx}{\textwidth }{ 
  >{\centering\arraybackslash}X >{\centering\arraybackslash}X >{\centering\arraybackslash}X >{\centering\arraybackslash}X >{\centering\arraybackslash}X >{\centering\arraybackslash}X 
   }
    Input & DWN~\cite{sharma2021wavelength}  & FUnIE-GAN~\cite{islam2020fast}  & WaterNet~\cite{li2019underwater} & Ucolor~\cite{Ucolor} & Ours \\ 
    \end{tabularx}
    \centering
    \includegraphics[width=\textwidth]{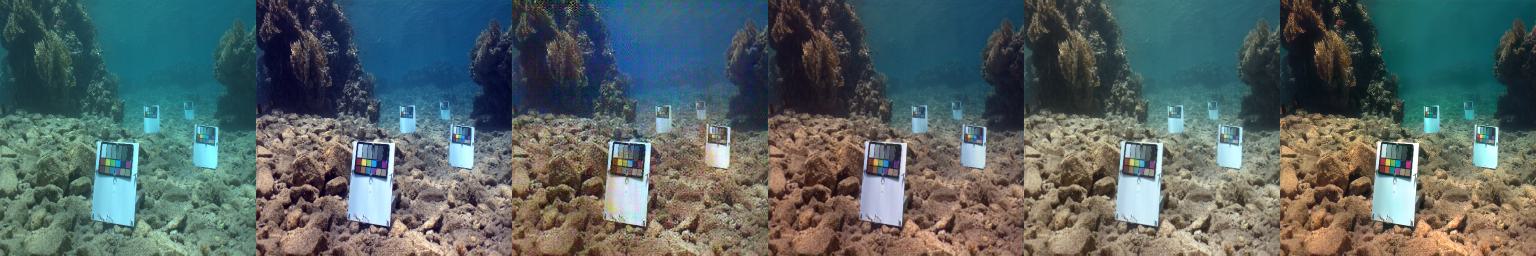}
        \begin{tabularx}{\textwidth }{ 
  >{\centering\arraybackslash}X >{\centering\arraybackslash}X >{\centering\arraybackslash}X >{\centering\arraybackslash}X >{\centering\arraybackslash}X >{\centering\arraybackslash}X 
   }
    Input & DWN~\cite{sharma2021wavelength}  & FUnIE-GAN~\cite{islam2020fast}  & WaterNet~\cite{li2019underwater} & Ucolor~\cite{Ucolor} & Ours \\ 
    \end{tabularx}
    \includegraphics[width=\textwidth]{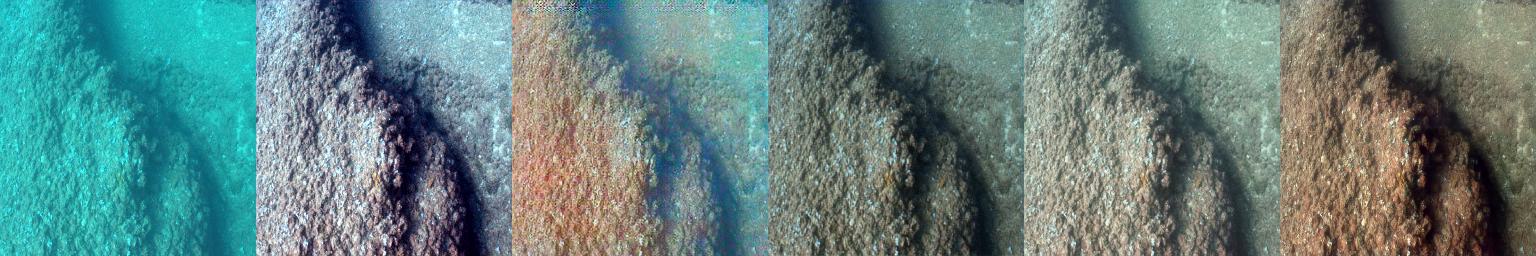}
        \begin{tabularx}{\textwidth }{ 
  >{\centering\arraybackslash}X >{\centering\arraybackslash}X >{\centering\arraybackslash}X >{\centering\arraybackslash}X >{\centering\arraybackslash}X >{\centering\arraybackslash}X 
   }
    Input & DWN~\cite{sharma2021wavelength}  & FUnIE-GAN~\cite{islam2020fast}  & WaterNet~\cite{li2019underwater} & Ucolor~\cite{Ucolor} & Ours \\ 
    \end{tabularx}
    \includegraphics[width=\textwidth]{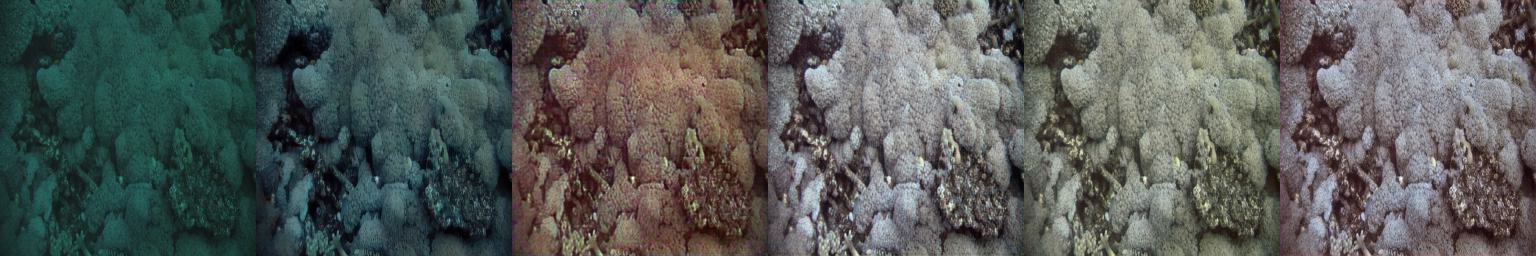}
            \begin{tabularx}{\textwidth }{ 
  >{\centering\arraybackslash}X >{\centering\arraybackslash}X >{\centering\arraybackslash}X >{\centering\arraybackslash}X >{\centering\arraybackslash}X >{\centering\arraybackslash}X 
   }
   Input & DWN~\cite{sharma2021wavelength}  & FUnIE-GAN~\cite{islam2020fast}  & WaterNet~\cite{li2019underwater} & Ucolor~\cite{Ucolor} & Ours \\ 
    \end{tabularx}
    \includegraphics[width=\textwidth]{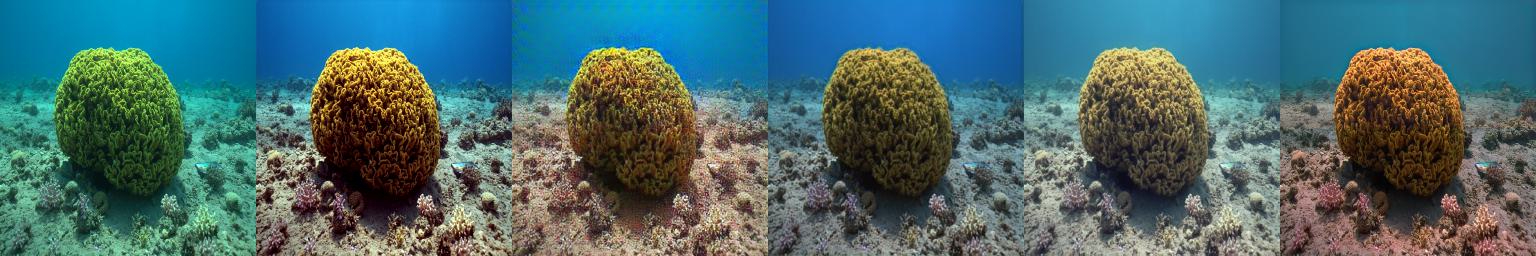}
            \begin{tabularx}{\textwidth }{ 
  >{\centering\arraybackslash}X >{\centering\arraybackslash}X >{\centering\arraybackslash}X >{\centering\arraybackslash}X >{\centering\arraybackslash}X >{\centering\arraybackslash}X 
   }
   Input & DWN~\cite{sharma2021wavelength}  & FUnIE-GAN~\cite{islam2020fast}  & WaterNet~\cite{li2019underwater} & Ucolor~\cite{Ucolor} & Ours \\ 
    \end{tabularx}
    \includegraphics[width=\textwidth]{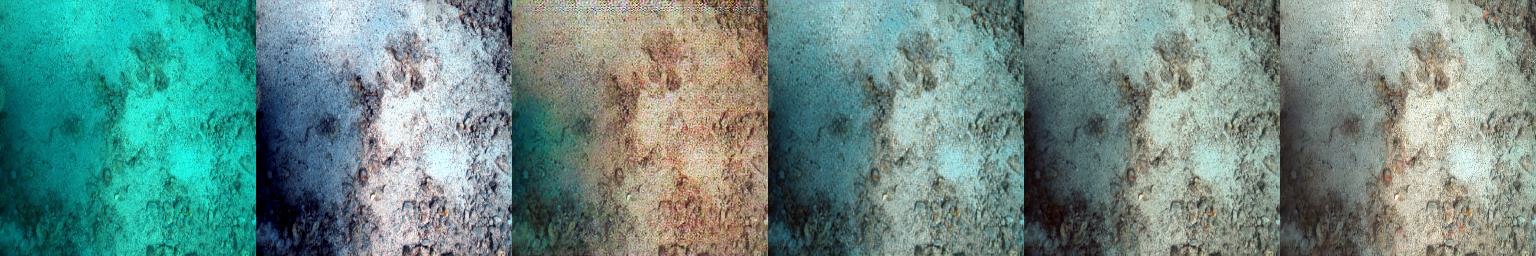}
            \begin{tabularx}{\textwidth }{ 
  >{\centering\arraybackslash}X >{\centering\arraybackslash}X >{\centering\arraybackslash}X >{\centering\arraybackslash}X >{\centering\arraybackslash}X >{\centering\arraybackslash}X 
   }
   Input & DWN~\cite{sharma2021wavelength}  & FUnIE-GAN~\cite{islam2020fast}  & WaterNet~\cite{li2019underwater} & Ucolor~\cite{Ucolor} & Ours \\ 
    \end{tabularx}
    \includegraphics[width=\textwidth]{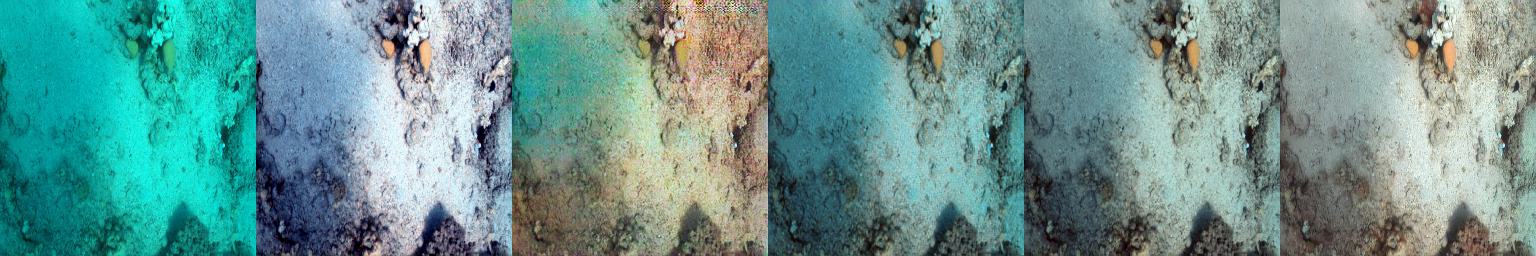}
            \begin{tabularx}{\textwidth }{ 
  >{\centering\arraybackslash}X >{\centering\arraybackslash}X >{\centering\arraybackslash}X >{\centering\arraybackslash}X >{\centering\arraybackslash}X >{\centering\arraybackslash}X 
   }
    \end{tabularx}
    \caption{Visual comparisons on real underwater images}
    \label{Comparisons_2}
\end{figure*}

\begin{figure*}[!t]
    \begin{tabularx}{\textwidth }{ 
  >{\centering\arraybackslash}X >{\centering\arraybackslash}X >{\centering\arraybackslash}X >{\centering\arraybackslash}X >{\centering\arraybackslash}X >{\centering\arraybackslash}X 
   }
    Input & DWN~\cite{sharma2021wavelength}  & FUnIE-GAN~\cite{islam2020fast}  & WaterNet~\cite{li2019underwater} & Ucolor~\cite{Ucolor} & Ours \\ 
    \end{tabularx}
    \centering
    \includegraphics[width=\textwidth]{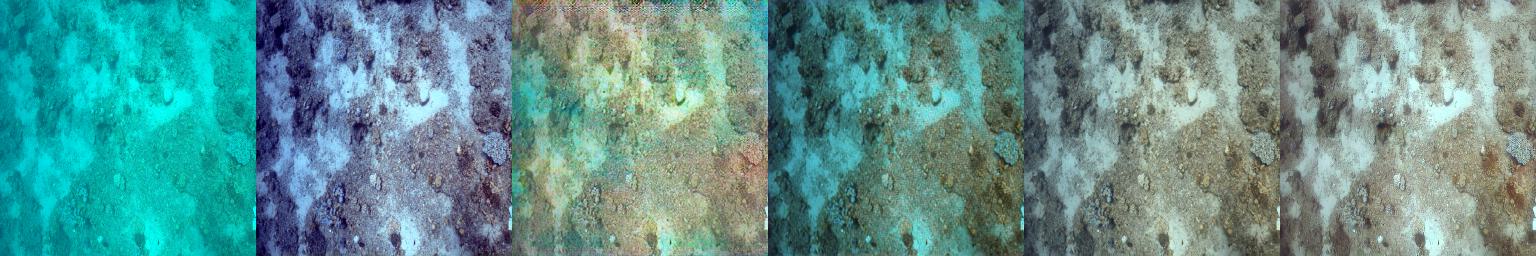}
        \begin{tabularx}{\textwidth }{ 
  >{\centering\arraybackslash}X >{\centering\arraybackslash}X >{\centering\arraybackslash}X >{\centering\arraybackslash}X >{\centering\arraybackslash}X >{\centering\arraybackslash}X 
   }
    Input & DWN~\cite{sharma2021wavelength}  & FUnIE-GAN~\cite{islam2020fast}  & WaterNet~\cite{li2019underwater} & Ucolor~\cite{Ucolor} & Ours \\ 
    \end{tabularx}
    \includegraphics[width=\textwidth]{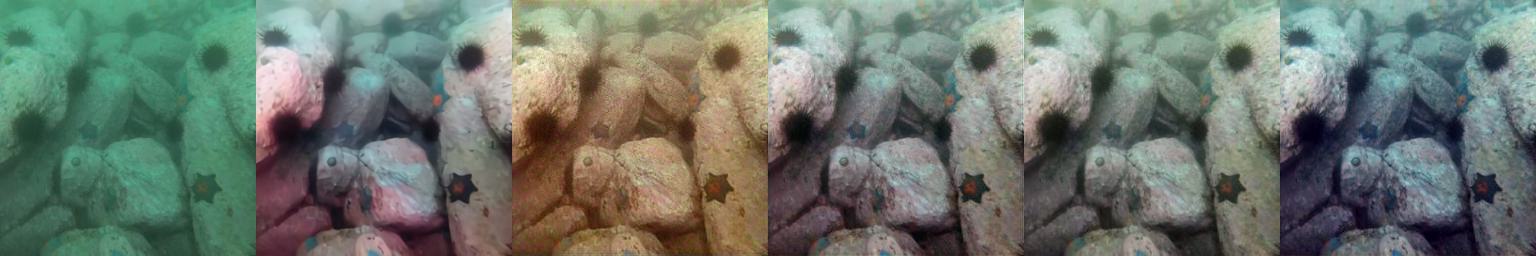}
        \begin{tabularx}{\textwidth }{ 
  >{\centering\arraybackslash}X >{\centering\arraybackslash}X >{\centering\arraybackslash}X >{\centering\arraybackslash}X >{\centering\arraybackslash}X >{\centering\arraybackslash}X 
   }
    Input & DWN~\cite{sharma2021wavelength}  & FUnIE-GAN~\cite{islam2020fast}  & WaterNet~\cite{li2019underwater} & Ucolor~\cite{Ucolor} & Ours \\ 
    \end{tabularx}
    \includegraphics[width=\textwidth]{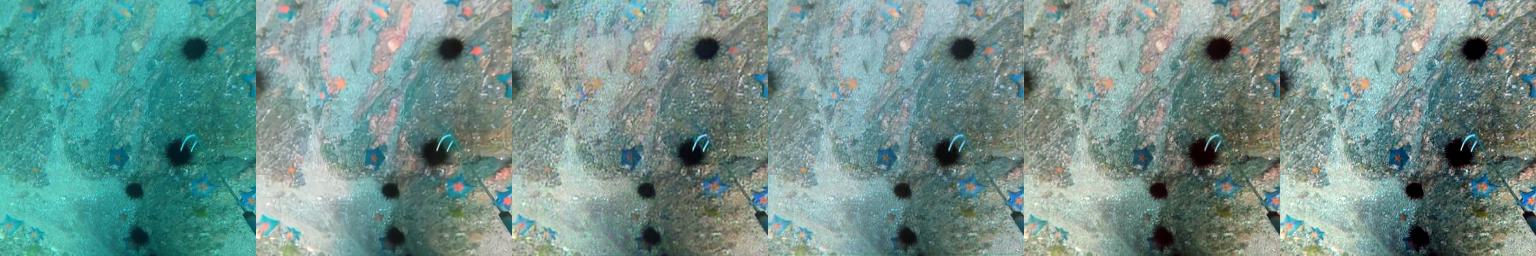}
            \begin{tabularx}{\textwidth }{ 
  >{\centering\arraybackslash}X >{\centering\arraybackslash}X >{\centering\arraybackslash}X >{\centering\arraybackslash}X >{\centering\arraybackslash}X >{\centering\arraybackslash}X 
   }
   Input & DWN~\cite{sharma2021wavelength}  & FUnIE-GAN~\cite{islam2020fast}  & WaterNet~\cite{li2019underwater} & Ucolor~\cite{Ucolor} & Ours \\ 
    \end{tabularx}
    \includegraphics[width=\textwidth]{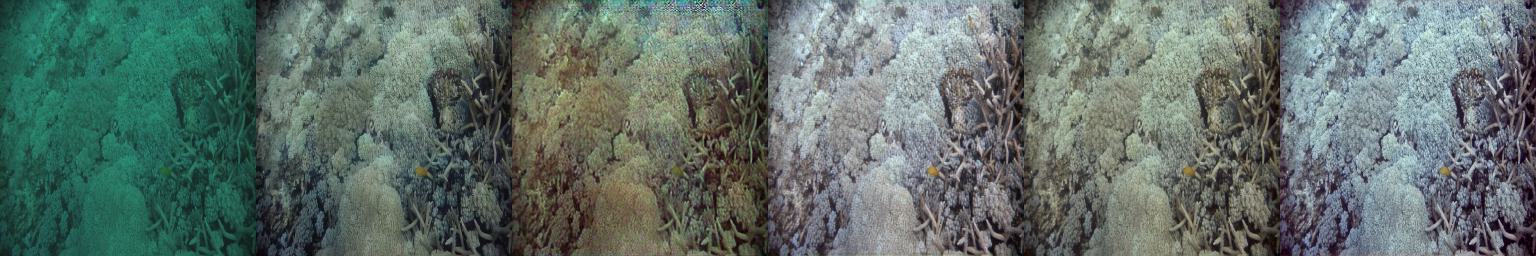}
            \begin{tabularx}{\textwidth }{ 
  >{\centering\arraybackslash}X >{\centering\arraybackslash}X >{\centering\arraybackslash}X >{\centering\arraybackslash}X >{\centering\arraybackslash}X >{\centering\arraybackslash}X 
   }
   Input & DWN~\cite{sharma2021wavelength}  & FUnIE-GAN~\cite{islam2020fast}  & WaterNet~\cite{li2019underwater} & Ucolor~\cite{Ucolor} & Ours \\ 
    \end{tabularx}
    \includegraphics[width=\textwidth]{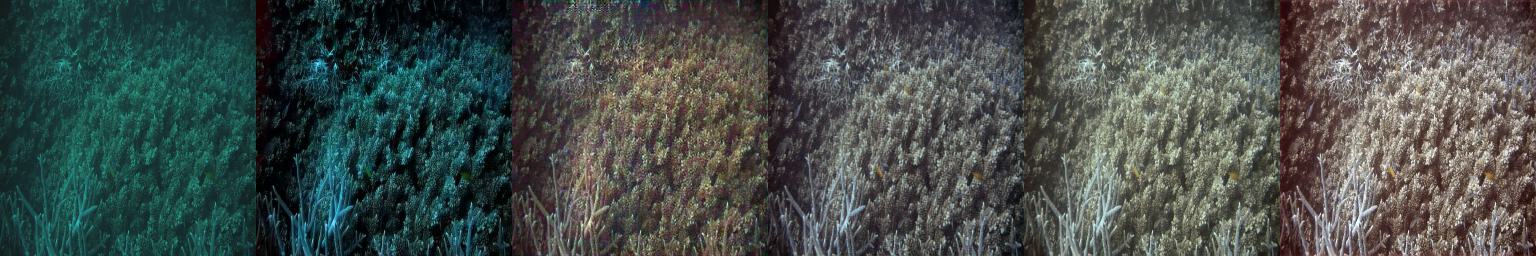}
            \begin{tabularx}{\textwidth }{ 
  >{\centering\arraybackslash}X >{\centering\arraybackslash}X >{\centering\arraybackslash}X >{\centering\arraybackslash}X >{\centering\arraybackslash}X >{\centering\arraybackslash}X 
   }
   Input & DWN~\cite{sharma2021wavelength}  & FUnIE-GAN~\cite{islam2020fast}  & WaterNet~\cite{li2019underwater} & Ucolor~\cite{Ucolor} & Ours \\ 
    \end{tabularx}
    \includegraphics[width=\textwidth]{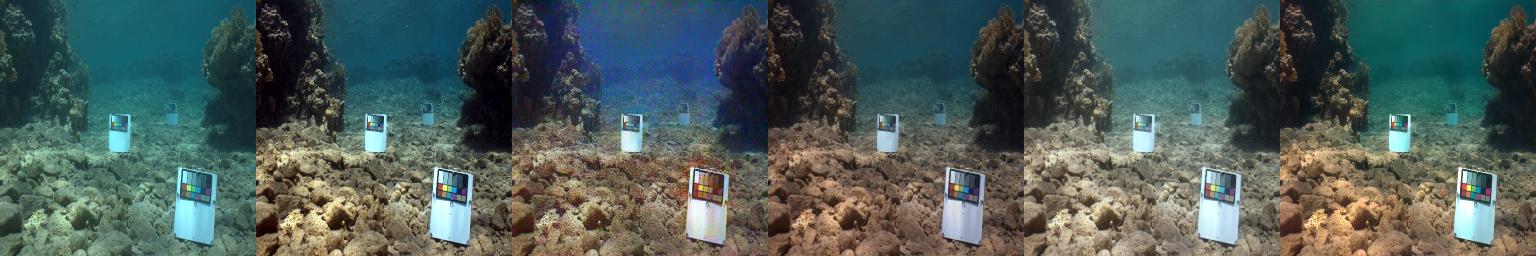}
            \begin{tabularx}{\textwidth }{ 
  >{\centering\arraybackslash}X >{\centering\arraybackslash}X >{\centering\arraybackslash}X >{\centering\arraybackslash}X >{\centering\arraybackslash}X >{\centering\arraybackslash}X 
   }
    \end{tabularx}
    \caption{More visual comparisons on real underwater images}
    \label{Comparisons_3}
\end{figure*}

We compare our method with existing state of the art deep methods \cite{sharma2021wavelength,islam2020fast,li2019underwater,Ucolor} for underwater restoration. In Fig. \ref{Comparisons}, we test our method on real underwater images collected by us during a diving expedition. Additionally, we test our method on other real underwater images of the SeaThru \cite{akkaynak2019sea} dataset, UIEBD \cite{li2019underwater} dataset and the HICRD dataset \cite{han2021cwr}. As reported in Fig. \ref{Comparisons_2} and Fig. \ref{Comparisons_3}, our method is able to remove the greenish/bluish hue from these underwater images. Thus, our restoration model is able to generalise to a wide variety of underwater conditions.

\subsection{Ablation on loss functions}

\begin{figure*}[!t]
    \begin{tabularx}{\textwidth }{ 
  >{\centering\arraybackslash}X >{\centering\arraybackslash}X >{\centering\arraybackslash}X >{\centering\arraybackslash}X
  >{\centering\arraybackslash}X
   }
    Synthetic underwater & Style Image & Baseline & Baseline+Edge & Ours \\ 
    \end{tabularx}
    \centering
    \includegraphics[width=\textwidth]{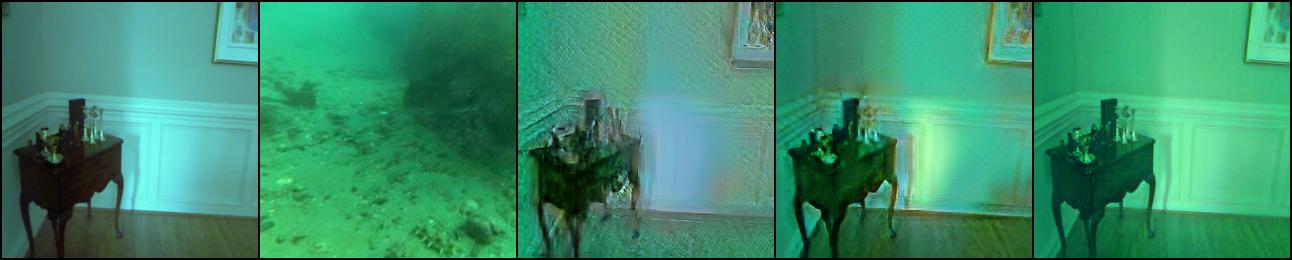}
    \includegraphics[width=\textwidth]{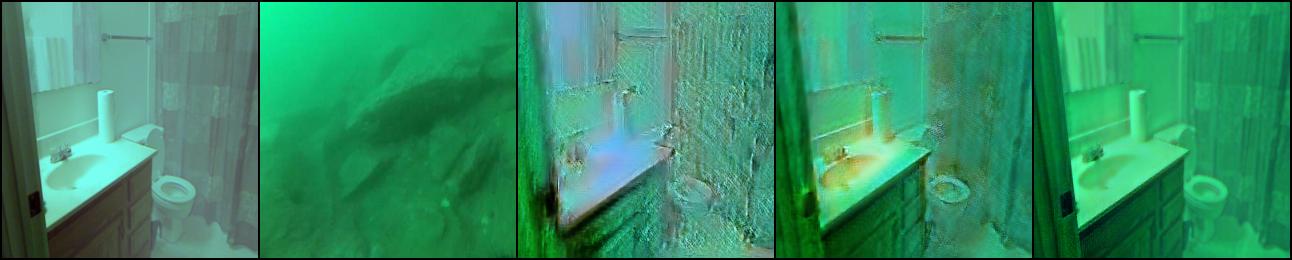}
    \includegraphics[width=\textwidth]{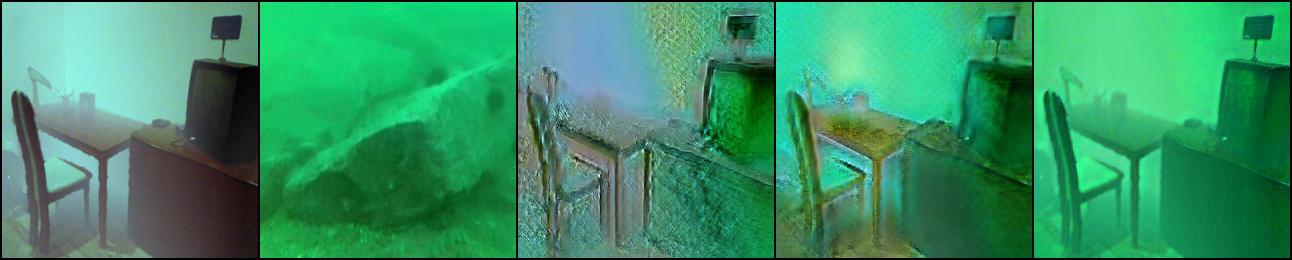}
    \caption{Results from our ablation study}
    \label{ablations}
\end{figure*}

We introduce additional losses such as SSIM loss and Edge loss to the domain adapted framework of \cite{huang2018multimodal}. We observed that the domain adapted dataset created using the  standard framework \cite{huang2018multimodal} had a lot of structural information missing and also had artifacts. To show the effectiveness of the losses introduced by us, we perform an ablation study. In this study, we consider the baseline results as the results obtained using just the standard losses of \cite{huang2018multimodal}. The 'Baseline+Edge' are the results obtained using the standard losses and using an additional edge loss (Eqn. 6). 'Ours' denotes the results obtained using the standard framework, edge loss (Eqn. 6) and SSIM Loss (Eqn. 7). The quantitative results of the ablation study can be found in Fig. \ref{ablations}.

\subsection{Some failure cases of our model}

\begin{figure*}[!t]
        \begin{tabularx}{\textwidth }{ 
  >{\centering\arraybackslash}X >{\centering\arraybackslash}X >{\centering\arraybackslash}X >{\centering\arraybackslash}X >{\centering\arraybackslash}X >{\centering\arraybackslash}X 
   }
    Input & DWN~\cite{sharma2021wavelength}  & FUnIE-GAN~\cite{islam2020fast}  & WaterNet~\cite{li2019underwater} & Ucolor~\cite{Ucolor} & Ours \\ 
    \end{tabularx}
    \includegraphics[width=\textwidth]{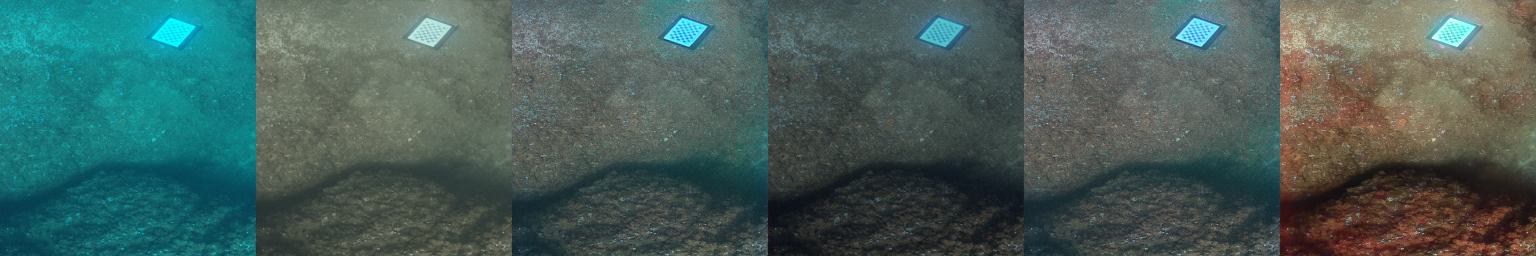}
        \begin{tabularx}{\textwidth }{ 
  >{\centering\arraybackslash}X >{\centering\arraybackslash}X >{\centering\arraybackslash}X >{\centering\arraybackslash}X >{\centering\arraybackslash}X >{\centering\arraybackslash}X 
   }
    Input & DWN~\cite{sharma2021wavelength}  & FUnIE-GAN~\cite{islam2020fast}  & WaterNet~\cite{li2019underwater} & Ucolor~\cite{Ucolor} & Ours \\ 
    \end{tabularx}
    \includegraphics[width=\textwidth]{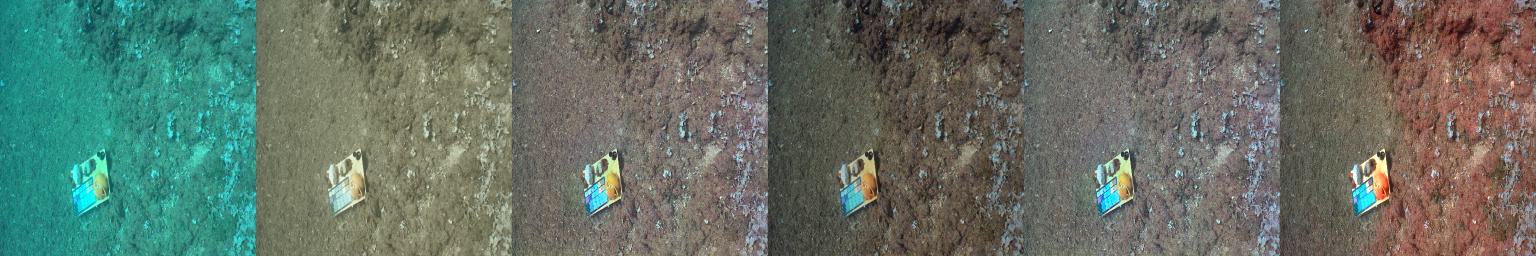}
            \begin{tabularx}{\textwidth }{ 
  >{\centering\arraybackslash}X >{\centering\arraybackslash}X >{\centering\arraybackslash}X >{\centering\arraybackslash}X >{\centering\arraybackslash}X >{\centering\arraybackslash}X 
   }
    \end{tabularx}
    \caption{A few failure cases for our model}
    \label{failure_cases}
\end{figure*}

Some images restored by our model have a reddish tinge. Underwater images generally have a bluish or greenish tinge due to significant attenuation of the red channel. Thus, the network needs to account for this by increasing the value of red channel. For some images, we hypothesize that our model tries to overcompensate this process  due to which some restored images have an overtly reddish tinge. 
A few failure cases can be found in Fig. \ref{failure_cases}.

\begin{center}
\end{center}

\begin{center}
\end{center}

\begin{center}
\end{center}

\begin{center}
\end{center}

\begin{center}
\end{center}

\bibliographystyle{ACM-Reference-Format}
\bibliography{supp}